\documentclass[10pt,twocolumn,letterpaper]{article}

\usepackage{cvpr}
\usepackage{times}
\usepackage{epsfig}
\usepackage{graphicx}
\usepackage{amsmath, bm}
\usepackage{amssymb}
\usepackage{multirow}
\usepackage[normalem]{ulem}
\useunder{\uline}{\ul}{}
\usepackage{subfigure}
\usepackage{color}

\usepackage[breaklinks=true,bookmarks=false]{hyperref}

\cvprfinalcopy 

\def\cvprPaperID{6240} 

\ifcvprfinal\fi
\begin{document}

\title{FBNet: Hardware-Aware Efficient ConvNet Design \\
via Differentiable Neural Architecture Search}

\author{Bichen Wu$^{1}$\thanks{Work done while interning at Facebook.}, Xiaoliang Dai$^{2*}$, Peizhao Zhang$^3$, Yanghan Wang$^3$, Fei Sun$^3$, \\
Yiming Wu$^3$, Yuandong Tian$^3$, Peter Vajda$^3$, Yangqing Jia$^3$, and Kurt Keutzer$^1$\\
$^1$UC Berkeley, $^2$Princeton University, $^3$Facebook Inc.\\
\tt\small \{bichen, keutzer\}@berkeley.edu, xdai@princeton.edu, \\
\tt\small \{stzpz, yanghan, feisun, wyiming, yuandong, vajdap, jiayq\}@fb.com
}

\maketitle

\begin{abstract}
Designing accurate and efficient ConvNets for mobile devices is challenging because the design space is combinatorially large. Due to this, previous neural architecture search (NAS) methods are computationally expensive. ConvNet architecture optimality depends on factors such as input resolution and target devices. However, existing approaches are too resource demanding for case-by-case redesigns. Also, previous work focuses primarily on reducing FLOPs, but FLOP count does not always reflect actual latency. To address these, we propose a differentiable neural architecture search (DNAS) framework that uses gradient-based methods to optimize ConvNet architectures, avoiding enumerating and training individual architectures separately as in previous methods. FBNets (\textit{\textbf{F}acebook-\textbf{B}erkeley-\textbf{Net}s}), a family of models discovered by DNAS surpass state-of-the-art models both designed manually and generated automatically. FBNet-B achieves 74.1\% top-1 accuracy on ImageNet with 295M FLOPs and 23.1 ms latency on a Samsung S8 phone, 2.4x smaller and 1.5x faster than MobileNetV2-1.3\cite{sandler2018mobilenetv2} with similar accuracy. Despite higher accuracy and lower latency than MnasNet\cite{tan2018mnasnet}, we estimate FBNet-B's search cost is 420x smaller than MnasNet's, at only 216 GPU-hours. Searched for different resolutions and channel sizes, FBNets achieve 1.5\% to 6.4\% higher accuracy than MobileNetV2. The smallest FBNet achieves 50.2\% accuracy and 2.9 ms latency (345 frames per second) on a Samsung S8. Over a Samsung-optimized FBNet, the iPhone-X-optimized model achieves a 1.4x speedup on an iPhone X. FBNet models are open-sourced at \url{https://github.com/facebookresearch/mobile-vision}.
\end{abstract}


\begin{figure}[h]
\begin{center}
\includegraphics[width=1.\linewidth]{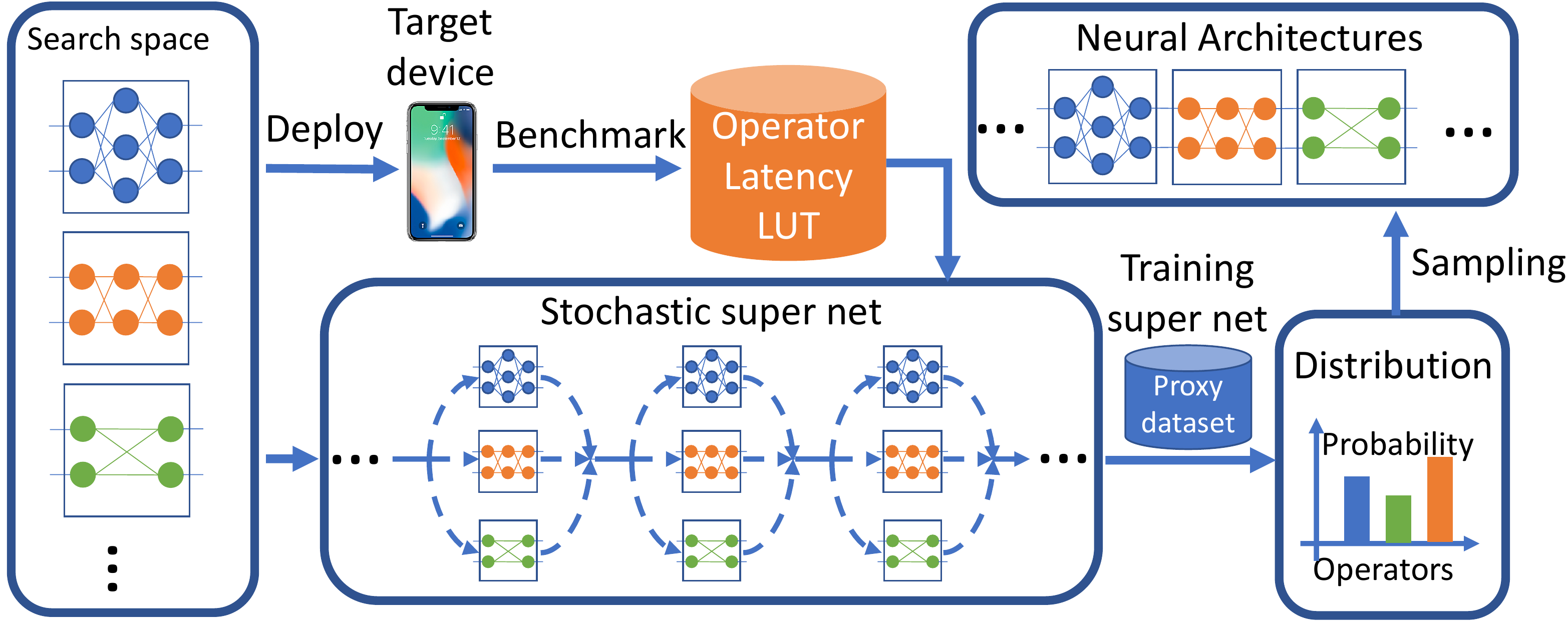}
\end{center}
\caption{Differentiable neural architecture search (DNAS) for ConvNet design. DNAS explores a layer-wise space that each layer of a ConvNet can choose a different block. The search space is represented by a stochastic super net. The search process trains the stochastic super net using SGD to optimize the architecture distribution. Optimal architectures are sampled from the trained distribution. The latency of each operator is measured on target devices and used to compute the loss for the super net.}
\label{fig:dnas_flow}
\end{figure}

\section{Introduction}
ConvNets are the \textit{de facto} method for computer vision. In many computer vision tasks, a better ConvNet design usually leads to significant accuracy improvement. In previous works, accuracy improvement comes at the cost of higher computational complexity, making it more challenging to deploy ConvNets to mobile devices, on which computing capacity is limited. Instead of solely focusing on accuracy, recent work also aims to optimize for efficiency, especially latency. However, designing efficient and accurate ConvNets is difficult due to the challenges below.

\begin{figure}[!t]
\begin{center}
\subfigure[A typical flow of manual ConvNet design.]{
\label{fig:manual_flow}
\includegraphics[width=0.9\linewidth]{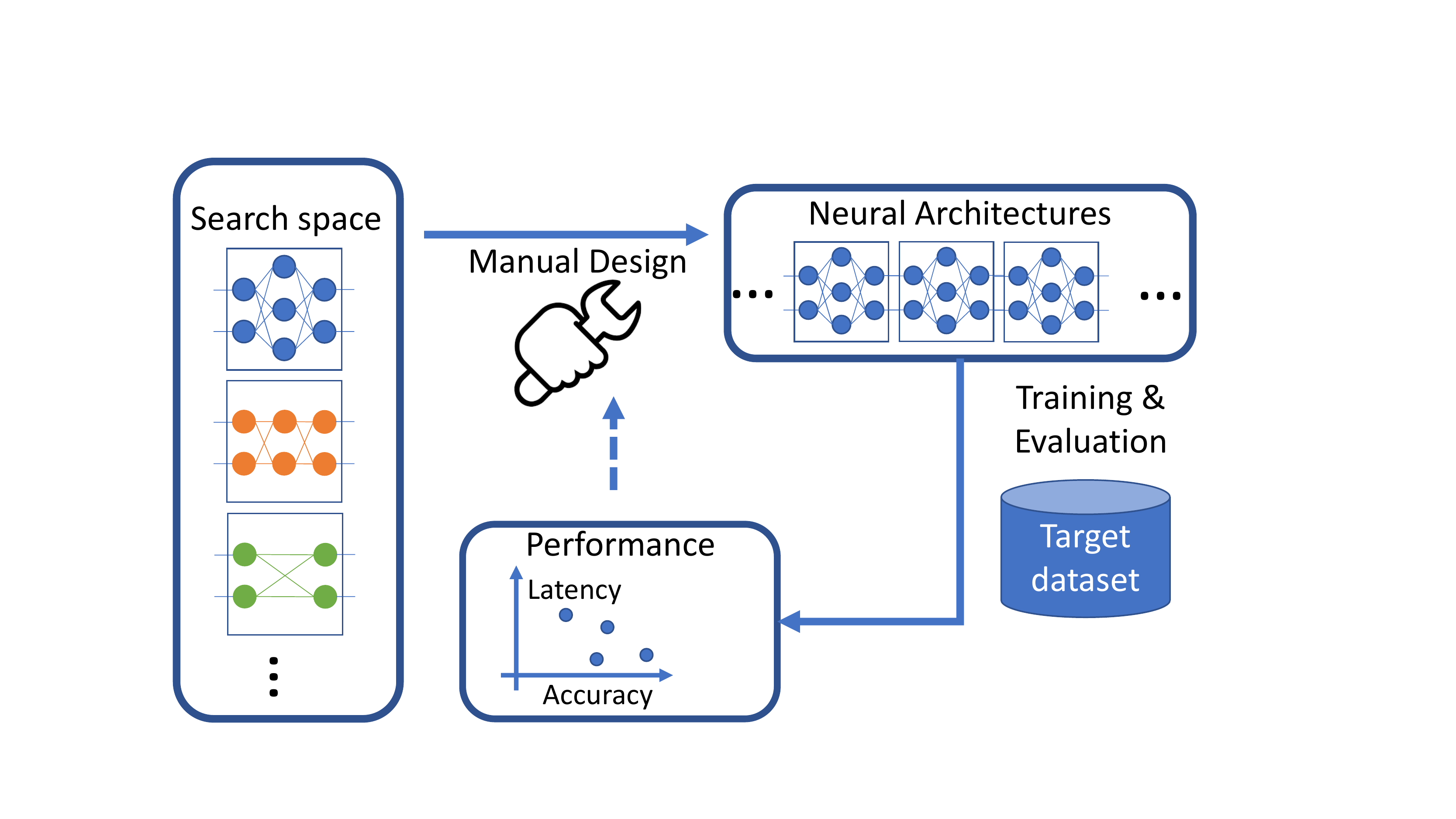}
}
\subfigure[A typical flow of reinforcement learning based neural architecture search.]{
\label{fig:rl_nas_flow}
\includegraphics[width=0.9\linewidth]{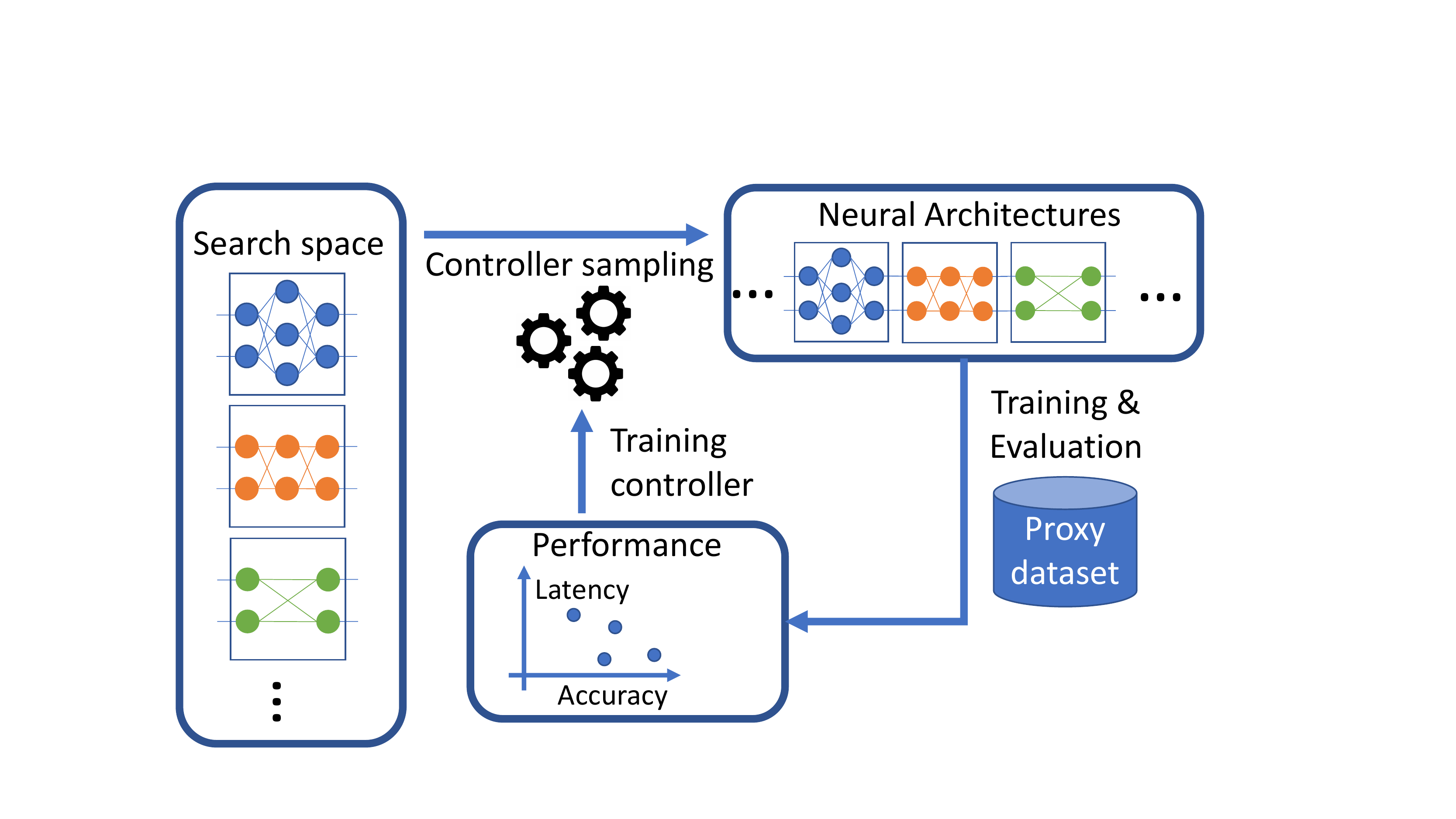}
}
\caption{Illustration of manual ConvNet design and reinforcement learning based neural architecture search.}
\label{fig:baseline_design_flow}
\end{center}
\vspace{-14pt}
\end{figure}

\textbf{Intractable design space}: The design space of a ConvNet is combinatorial. Using VGG16 \cite{simonyan2014very} as a motivating example: VGG16 contains 16 layers. Assume for each layer of the network, we can choose a different kernel size from $\{1, 3, 5\}$ and a different filter number from $\{32, 64, 128, 256, 512\}$. Even with such simplified design choices and shallow layers, the design space contains $(3\times 5)^{16} \approx 6\times 10^{18}$ possible architectures. However, training a ConvNet is very time-consuming, typically taking days or even weeks. As a result, previous ConvNet design rarely explores the design space. A typical flow of manual ConvNet design is illustrated in Figure \ref{fig:manual_flow}. Designers propose initial architectures and train them on the target dataset. Based on the performance, designers evolve the architectures accordingly. Limited by the time cost of training ConvNets, the design flow has to stop after a few iterations, which is far too few to sufficiently explore the design space. 

Starting from \cite{zoph2016neural}, recent works adopt neural architecture search (NAS) to explore the design space automatically. Many previous works \cite{zoph2016neural,zoph2017learning, tan2018mnasnet} use reinforcement learning (RL) to guide the search and a typical flow is illustrated in Figure \ref{fig:rl_nas_flow}. A controller samples architectures from the search space to be trained. To reduce the training cost, sampled architectures are trained on a smaller proxy dataset such as CIFAR-10 or trained for fewer epochs on ImageNet. The performance of the trained networks is then used to train and improve the controller. Previous works \cite{zoph2016neural,zoph2017learning, tan2018mnasnet} has demonstrated the effectiveness of such methods in finding accurate and efficient ConvNet models. However, training each architecture is still time-consuming, and it usually takes thousands of architectures to train the controller. As a result, the computational cost of such methods is prohibitively high.

\textbf{Nontransferable optimality}: the optimality of ConvNet architectures is conditioned on many factors such as input resolutions and target devices. Once these factors change, the optimal architecture is likely to be different. A common practice to reduce the FLOP count of a network is to shrink the input resolution. A smaller input resolution may require a smaller receptive field of the network and therefore shallower layers. On a different device, the same operator can have different latency, so we need to adjust the ConvNet architecture to achieve the best accuracy-efficiency trade-off. Ideally, we should design different ConvNet architectures case-by-case. In practice, however, limited by the computational cost of previous manual and automated approaches, we can only realistically design one ConvNet and use it for all conditions.

\textbf{Inconsistent efficiency metrics}: Most of the efficiency metrics we care about are dependent on not only the ConvNet architecture but also the hardware and software configurations on the target device. Such metrics include latency, power, energy, and in this paper, we mainly focus on latency. To simplify the problem, most of the previous works adopt hardware-agnostic metrics such as FLOPs (more strictly, number of multiply-add operations) to evaluate a ConvNet's efficiency. However, a ConvNet with lower FLOP count is not necessarily faster. For example, NasNet-A \cite{zoph2017learning} has a similar FLOP count as MobileNetV1 \cite{howard2017mobilenets}, but its complicated and fragmented cell-level structure is not hardware friendly, so the actual latency is slower \cite{sandler2018mobilenetv2}. The inconsistency between hardware agnostic metrics and actual efficiency makes the ConvNet design more difficult. 

To address the above problems, we propose to use differentiable neural architecture search (DNAS) to discover hardware-aware efficient ConvNets. The flow of our algorithm is illustrated in Figure \ref{fig:dnas_flow}. DNAS allows us to explore a layer-wise search space where we can choose a different block for each layer of the network. Following \cite{veniat2017learning}, DNAS represents the search space by a super net whose operators execute stochastically. We relax the problem of finding the optimal architecture to find a distribution that yields the optimal architecture. By using the Gumbel Softmax technique \cite{jang2016categorical}, we can directly train the architecture distribution using gradient-based optimization such as SGD. The search process is extremely fast compared with previous reinforcement learning (RL) based method. The loss used to train the stochastic super net consists of both the cross-entropy loss that leads to better accuracy and the latency loss that penalizes the network's latency on a target device. To estimate the latency of an architecture, we measure the latency of each operator in the search space and use a lookup table model to compute the overall latency by adding up the latency of each operator. Using this model allows us to quickly estimate the latency of architectures in this enormous search space. More importantly, it makes the latency differentiable with respect to layer-wise block choices. 

We name the models discovered by DNAS as FBNets (\textit{\textbf{F}acebook-\textbf{B}erkeley-\textbf{Net}s}). FBNets surpass the state-of-the-art efficient ConvNets designed manually and automatically. FBNet-B achieves 74.1\% top-1 accuracy with 295M FLOPs and 23.1 ms latency on an Samsung S8 phone, 2.4x smaller and 1.5x faster than MobileNetV2-1.3. Being better than MnasNet, FBNet-B's search cost is 216 GPU-hours, 421x lower than the cost for MnasNet estimated based on \cite{tan2018mnasnet}. Such low search cost enables us to re-design ConvNets case-by-case. For different resolution and channel scaling, FBNets achieve 1.5\% to 6.4\% absolute gain in top-1 accuracy compared with MobileNetV2 models. The smallest FBNet achieves 50.2\% accuracy and 2.9 ms latency (345 frames per second) with a batch size of 1 on Samsung S8. Using DNAS to search for device-specific ConvNet, an iPhone-x-optimized model achieves 1.4x speedup on an iPhone X compared with a Samsung-optimized model.

\section{Related work}
\textbf{Efficient ConvNet models}: Designing efficient ConvNet has attracted many research attention in recent years. SqueezeNet \cite{iandola2016squeezenet} is one of the early works focusing on reducing the parameter size of ConvNet models. It is originally designed for classification, but later extended to object detection \cite{wu2017squeezedet} and LiDAR point-cloud segmentation \cite{wu2018squeezeseg, wu2018squeezesegv2}. Following SqueezeNet, SqueezeNext \cite{gholami2018squeezenext} and ShiftNet \cite{wu2017shift} achieve further parameter size reduction. Recent works change the focus from parameter size to FLOPs. MobileNetV1 and MobileNetV2 \cite{howard2017mobilenets, sandler2018mobilenetv2} use depthwise convolutions to replace the more expensive spatial convolutions. ShuffleNet \cite{zhang1707shufflenet} uses group convolution and shuffle operations to reduce the FLOP count further. More recent works realize that FLOP count does not always reflect the actual hardware efficiency. To improve actual latency, ShuffleNetV2 \cite{ma2018shufflenet} proposes a series of practical guidelines for efficient ConvNet design. Synetgy \cite{yang2018synetgy} combines ideas from ShuffleNetV2 and ShiftNet to co-design hardware friendly ConvNets and FPGA accelerators. 

\textbf{Neural Architecture Search}: \cite{zoph2016neural, zoph2017learning} first proposes to use reinforcement learning (RL) to search for neural architectures to achieve competitive accuracy with low FLOPs. Early NAS methods are computationally expensive. Recent works try to reduce the computational cost by weight sharing \cite{pham2018efficient} or using gradient-based optimization \cite{liu2018darts}. \cite{wu2018mixed, anonymous2019snas:} further develop the idea of differentiable neural architecture search combining Gumbel Softmax \cite{jang2016categorical}. Early works of NAS \cite{zoph2017learning,pham2018efficient,liu2018darts} focus on the cell level architecture search, and the same cell structure is repeated in all layers of a network. However, such fragmented and complicated cell-level structures are not hardware friendly, and the actual efficiency is low. Most recently, \cite{tan2018mnasnet} explores a stage-level hierarchical search space, allowing different blocks for different stages of a network, while blocks inside a stage are still the same. Instead of focusing on FLOPs, \cite{tan2018mnasnet} aims to optimize the latency on target devices.  Besides searching for new architectures, works such as \cite{yang2018netadapt, he2018amc} focus on adapting existing models to improve efficiency.

\section{Method}
In this paper, we use differentiable neural architecture search (DNAS) to solve the problem of ConvNet design. We formulate the neural architecture search problem as
\begin{equation}
\label{eqn:nas}
\underset{a \in \mathcal{A}}{\text{min }} \underset{w_a}{\text{min }} \mathcal{L}(a, w_a).
\end{equation}
Given an architecture space $\mathcal{A}$, we seek to find an optimal architecture $a \in \mathcal{A}$ such that after training its weights $w_a$, it can achieve the minimal loss $\mathcal{L}(a, w_a)$. In our work, we focus on three factors of the problem: a) the search space $\mathcal{A}$. b) The loss function $\mathcal{L}(a, w_a)$ that considers actual latency. c) An efficient search algorithm.

\subsection{The Search Space}
\label{sec:search_space}
Previous works \cite{zoph2016neural,zoph2017learning,pham2018efficient,liu2017progressive,liu2018darts} focus on cell level architecture search. Once a cell structure is searched, it is used in all the layers across the network. However, many searched cell structures are very complicated and fragmented and are therefore slow when deployed to mobile CPUs \cite{sandler2018mobilenetv2,ma2018shufflenet}. Besides, at different layers, the same cell structure can have a different impact on the accuracy and latency of the overall network. As shown in \cite{tan2018mnasnet} and in our experiments, allowing different layers to choose different blocks leads to better accuracy and efficiency. 

In this work, we construct a layer-wise search space with a fixed macro-architecture, and each layer can choose a different block. The macro-architecture is described in Table \ref{tab:macro-space}. The macro architecture defines the number of layers and the input/output dimensions of each layer. The first and the last three layers of the network have fixed operators. For the rest of the layers, their block type needs to be searched. The filter numbers for each layer are hand-picked empirically. We use relatively small channel sizes for early layers, since the input resolution at early layers is large, and the computational cost (FLOP count) is quadratic to input size. 

\begin{table}[h]
\centering
\begin{tabular}{c|c|c|c|c}
\hline
Input shape             & Block       & f         & n        & s \\ \hline
$224^2 \times 3$        & 3x3 conv    & 16        & 1        & 2      \\
$112^2 \times 16$       & TBS         & 16        & 1        & 1      \\
$112^2 \times 16$       & TBS         & 24        & 4        & 2      \\
$56^2 \times 24$        & TBS         & 32        & 4        & 2      \\
$28^2 \times 32$        & TBS         & 64        & 4        & 2      \\
$14^2 \times 64$        & TBS         & 112       & 4        & 1      \\
$14^2 \times 112$       & TBS         & 184       & 4        & 2      \\
$7^2 \times 184$        & TBS         & 352       & 1        & 1      \\
$7^2 \times 352$        & 1x1 conv    & 1504 (1984)      & 1        & 1      \\
$7^2 \times 1504~(1984)$       & 7x7 avgpool & -         & 1        & 1      \\
$1504$                  & fc          & 1000      & 1        & -      \\ \hline
\end{tabular}
\caption{Macro-architecture of the search space. Column-``Block'' denotes the block type. ``TBS'' means layer type needs to be searched. Column-$f$ denotes the filter number of a block. Column-$n$ denotes the number of blocks. Column-$s$ denotes the stride of the first block in a stage.  The filter size of the last 1x1 conv is 1504 for FBNet-A and 1984 for FBNet-\{B, C\}.}
\label{tab:macro-space}
\end{table}

\begin{figure}[!t]
\begin{center}
\centering \includegraphics[width=0.5\linewidth]{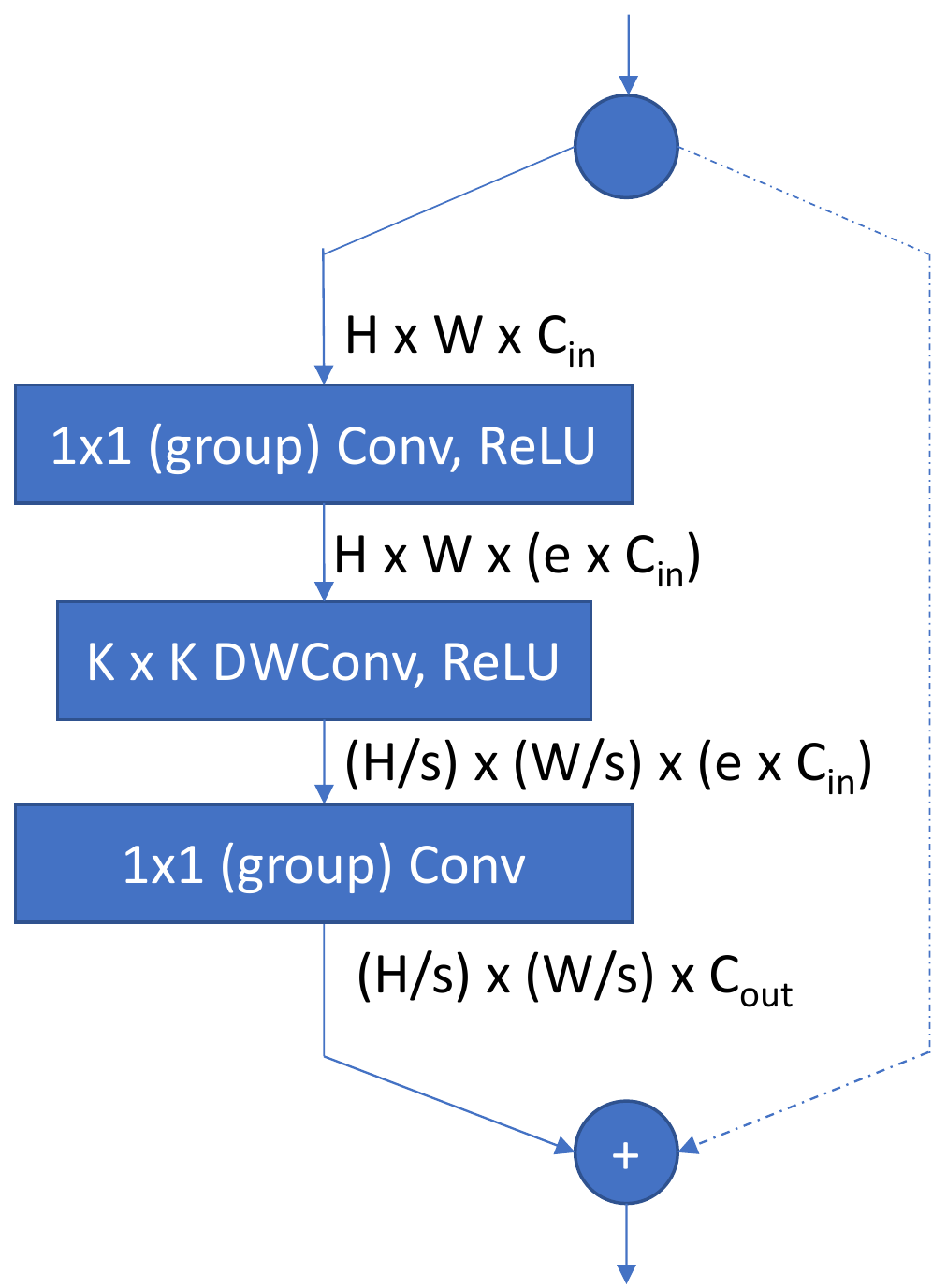}
\caption{The block structure of the micro-architecture search space. Each candidate block in the search space can choose a different expansion rate, kernel size, and number of groups for group convolution.}
\label{fig:block}
\end{center}
\vspace{-20pt}
\end{figure}

Each searchable layer in the network can choose a different block from the layer-wise search space. The block structure is inspired by MobileNetV2 \cite{sandler2018mobilenetv2} and ShiftNet \cite{wu2017shift}, and is illustrated in Figure \ref{fig:block}. It contains a point-wise (1x1) convolution, a K-by-K depthwise convolution where K denotes the kernel size, and another 1x1 convolution. ``ReLU'' activation functions follow the first 1x1 convolution and the depthwise convolution, but there are no activation functions following the last 1x1 convolution. If the output dimension stays the same as the input dimension, we use a skip connection to add the input to the output. Following \cite{sandler2018mobilenetv2, wu2017shift}, we use a hyperparameter, the expansion ratio $e$, to control the block. It determines how much do we expand the output channel size of the first 1x1 convolution compared with its input channel size. Following \cite{tan2018mnasnet}, we also allow choosing a kernel size of 3 or 5 for the depthwise convolution. In addition, we can choose to use group convolution for the first and the last 1x1 convolution to reduce the computation complexity. When we use group convolution, we follow \cite{zhang1707shufflenet} to add a channel shuffle operation to mix the information between channel groups. 

In our experiments, our layer-wise search space contains 9 candidate blocks, with their configurations listed in Table \ref{tab:micro-space}. Note we also have a block called ``skip'', which directly feed the input feature map to the output without actual computations. This candidate block essentially allows us to reduce the depth of the network. 

In summary, our overall search space contains 22 layers and each layer can choose from 9 candidate blocks from Table \ref{tab:micro-space}, so it contains $9^{22} \approx 10^{21}$ possible architectures. Finding the optimal layer-wise block assignment from such enormous search space is a non-trivial task.

\begin{table}[]
\centering
\begin{tabular}{c|c|c|c}
\hline
Block type & expansion      & Kernel      & Group \\ \hline
k3\_e1     & 1              & 3           & 1     \\
k3\_e1\_g2 & 1              & 3           & 2     \\
k3\_e3     & 3              & 3           & 1     \\
k3\_e6     & 6              & 3           & 1     \\
k5\_e1     & 1              & 5           & 1     \\
k5\_e1\_g2 & 1              & 5           & 2     \\
k5\_e3     & 3              & 5           & 1     \\
k5\_e6     & 6              & 5           & 1     \\ 
skip       & -              & -           & -     \\ \hline
\end{tabular}
\caption{Configurations of candidate blocks in the search space.}
\label{tab:micro-space}
\end{table}

\subsection{Latency-Aware Loss Function}
\label{sec:loss_function}
The loss function used in (\ref{eqn:nas}) has to reflect not only the accuracy of a given architecture but also the latency on the target hardware. To achieve this goal, we define the following loss function:
\begin{equation}
\label{eqn:loss}
    \mathcal{L}(a, w_a) = \text{ CE}(a, w_a) \cdot \alpha \log(\text{LAT}(a))^\beta.
\end{equation}
The first term $\text{CE}(a, w_a)$ denotes the cross-entropy loss of architecture $a$ with parameter $w_a$. The second term $\text{LAT}(a)$ denotes the latency of the architecture on the target hardware measured in micro-second. The coefficient $\alpha$ controls the overall magnitude of the loss function. The exponent coefficient $\beta$ modulates the magnitude of the latency term. 

The cross-entropy term can be easily computed. However, the latency term is more difficult, since we need to measure the actual runtime of an architecture on a target device. To cover the entire search space, we need to measure about $10^{21}$ architectures, which is an impossible task. 

To solve this problem, we use a latency lookup table model to estimate the overall latency of a network based on the runtime of each operator. More formally, we assume
\begin{equation}
\label{eqn:latency}
    \text{LAT}(a) = \sum_l \text{LAT} (b_l^{(a)}),
\end{equation}
where $b_l^{(a)}$ denotes the block at layer-$l$ from architecture $a$. This assumes that on the target processor, the runtime of each operator is independent of other operators. The assumption is valid for many mobile CPUs and DSPs, where operators are computed sequentially one by one. This way, by benchmarking the latency of a few hundred operators used in the search space, we can easily estimate the actual runtime of the $10^{21}$ architectures in the entire search space. More importantly, as will be explained in section \ref{sec:dnas}, using the lookup table model makes the latency term in the loss function (\ref{eqn:loss}) differentiable with respect to layer-wise block choices, and this allows us to use gradient-based optimization to solve problem (\ref{eqn:nas}). 

\subsection{The Search Algorithm}
\label{sec:dnas}
Solving the problem (\ref{eqn:nas}) through brute-force enumeration of the search space is very infeasible. The inner problem of optimizing $w_a$ involves training a neural network. For ImageNet classification, training a ConvNet typically takes several days or even weeks. The outer problem of optimizing $a \in \mathcal{A}$ has a combinatorially large search space. 

Most of the early works on NAS \cite{zoph2016neural, zoph2017learning, tan2018mnasnet} follow the paradigm above. To reduce the computational cost, the inner problem is replaced by training candidate architectures on an easier proxy dataset. For example, \cite{zoph2016neural, zoph2017learning} trains the architecture on the CIFAR10 dataset, and \cite{tan2018mnasnet} trains on ImageNet but only for 5 epochs. The learned architectures are then transferred to the target dataset. To avoid exhaustively iterating through the search space, \cite{zoph2016neural,zoph2017learning,tan2018mnasnet} use reinforcement learning to guide the exploration. Despite these improvements, solving problem (\ref{eqn:nas}) is still prohibitively expensive -- training a network on the proxy dataset is still time-consuming, and thousands of architectures need to be trained before reaching the optimal solution.

We adopt a different paradigm of solving problem (\ref{eqn:nas}). We first represent the search space by a stochastic super net. The super net has the same macro-architecture as described in Table \ref{tab:macro-space}, and each layer contains 9 parallel blocks as described in Table \ref{tab:micro-space}.  During the inference of the super net, only one candidate block is sampled and executed with the sampling probability of
\begin{equation}
\label{eqn:softmax_prob}
    P_{\bm{\theta}_{l}}(b_l = b_{l,i}) = \text{softmax}(\theta_{l,i}; \bm{\theta}_l) = \frac{\exp(\theta_{l,i})}{\sum_i \exp(\theta_{l,i})}. 
\end{equation}
$\bm{\theta}_l$ contains parameters that determine the sampling probability of each block at layer-$l$. Equivalently, the output of layer-$l$ can be expressed as
\begin{equation}
\label{eqn:mask}
    x_{l+1} = \sum_i m_{l, i} \cdot b_{l, i}(x_{l}),
\end{equation}
where $m_{l,i}$ is a random variable in $\{0, 1\}$ and is evaluated to 1 if block $b_{l, i}$ is sampled. The sampling probability is determined by equation (\ref{eqn:softmax_prob}). $b_{l, i}(x_{l})$ denotes the output of block-$i$ at layer $l$ given the input feature map $x_{l}$. We let each layer sample independently, therefore, the probability of sampling an architecture $a$ can be described as
\begin{equation}
\label{eqn:prob}
    P_{\bm{\theta}}(a) = \prod_l P_{\bm{\theta}_l} (b_l = b_{l,i}^{(a)}),
\end{equation}
where $\bm{\theta}$ denotes the a vector consists of all the $\theta_{l,i}$ for each block-$i$ at layer-$l$. $b_{l,i}^{(a)}$ denotes that in the sampled architecture $a$,  block-$i$ is chosen at layer-$l$. 

Instead of solving for the optimal architecture $a \in \mathcal{A}$, which has a discrete search space, we relax the problem to optimize the probability $P_\theta$ of the stochastic super net to achieve the minimum expected loss. Formally, we re-write the discrete optimization problem (\ref{eqn:nas}) as 
\begin{equation}
\label{eqn:dnas}
\underset{\bm{\theta}}{\text{min }} \underset{w_a}{\text{min }} \mathbf{E}_{a \sim P_{\bm{\theta}}} \{\mathcal{L}(a, w_a) \}.
\end{equation}
It is obvious the loss function in (\ref{eqn:dnas}) is differentiable with respect to the architecture weights $w_a$ and therefore can be optimized by stochastic gradient descent (SGD). However, the loss is not directly differentiable to the sampling parameter $\bm{\theta}$, since we cannot pass the gradient through the discrete random variable $m_{l,i}$ to $\theta_{l,i}$. To sidestep this, we relax the discrete mask variable $m_{l,i}$ to be a continuous random variable computed by the Gumbel Softmax function \cite{jang2016categorical, maddison2016concrete}
\begin{equation}
\label{eqn:gumbel_softmax}
\begin{aligned}
    m_{l, i} & = \text{GumbelSoftmax}(\theta_{l, i}|\bm{\theta_{l}}) \\
             & = \frac{\exp[(\theta_{l,i} + g_{l,i})/\tau]}{\sum_i \exp[(\theta_{l,i} + g_{l,i})/\tau]}, 
\end{aligned}
\end{equation}
where $g_{l,i} \sim \text{Gumbel(0, 1)}$ is a random noise following the Gumbel distribution. The Gumbel Softmax function is controlled by a temperature parameter $\tau$. As $\tau$ approaches 0, it approximates the discrete categorical sampling following the distribution in (\ref{eqn:prob}). As $\tau$ becomes larger, $m_{l,i}$ becomes a continuous random variable. Regardless of the value of $\tau$, the mask $m_{l,i}$ is directly differentiable with respect to the parameter $\theta_{l, i}$. The technique of using Gumbel Softmax for neural architecture search is also proposed in \cite{wu2018mixed, anonymous2019snas:}.

As a result, it is clear that the cross-entropy term from the loss function (\ref{eqn:loss}) is differentiable with respect to the mask $m_{l,i}$ and therefore $\theta_{l,i}$. For the latency term, since we use the lookup table based model for efficiency estimation, equation (\ref{eqn:latency}) can be written as
\begin{equation}
\label{eqn:lut-latency}
    \text{LAT}(a) = \sum_l \sum_i m_{l,i} \cdot \text{LAT} (b_{l,i}).
\end{equation}
The latency of each operator $\text{LAT} (b_{l,i})$ is a constant coefficient, so the overall latency of architecture-$a$ is differentiable with respect to the mask $m_{l,i}$, therefore $\theta_{l,i}$. 

As a result, the loss function (\ref{eqn:loss}) is fully differentiable with respect to both weights $w_a$ and the architecture distribution parameter $\bm{\theta}$. This allows us to use SGD to efficiently solve problem (\ref{eqn:nas}). 

Our search process is now equivalent to training the stochastic super net. During the training, we compute $\partial \mathcal{L} / \partial w_a$ to train each operator's weight in the super net. This is no different from training an ordinary ConvNet. After operators get trained, different operators can have a different contribution to the accuracy and the efficiency of the overall network. Therefore, we compute $\partial \mathcal{L}/\partial \bm{\theta}$ to update the sampling probability $P_{\bm{\theta}}$ for each operator. This step selects operators with better accuracy and lower latency and suppresses the opposite ones. After the super net training finishes, we can then obtain the optimal architectures by sampling from the architecture distribution $P_{\bm{\theta}}$. 

As will be shown in the experiment section, the proposed DNAS algorithm is orders of magnitude faster than previous RL based NAS while generating better architectures. 

\section{Experiments}
\subsection{ImageNet Classification}
To demonstrate the efficacy of our proposed method, we use DNAS to search for ConvNet models on ImageNet 2012 classification dataset \cite{deng2009imagenet}, and we name the discovered models FBNets. We aim to discover models with high accuracy and low latency on target devices. In our first experiment, we target Samsung Galaxy S8 with a Qualcomm Snapdragon 835 platform. The model is deployed with Caffe2 with int8 inference engine for mobile devices.

Before the search starts, we first build a latency lookup table described in section \ref{sec:loss_function} on the target device. Next, we train a stochastic super net with search space described in section \ref{sec:dnas}. We set the input resolution of the network to 224-by-224. To reduce the training time, we randomly choose 100 classes from the original 1000 classes to train the stochastic super net. We train the stochastic super net for 90 epochs. In each epoch, we first train the operator weights $w_a$ and then the architecture probability parameter $\bm{\theta}$. $w_a$ is trained on 80\% of ImageNet training set using SGD with momentum. 
The architecture distribution parameter $\bm{\theta}$ is trained on the rest 20\% of ImageNet training set with Adam optimizer \cite{kingma2014adam}.
To control the temperature of the Gumbel Softmax from equation (\ref{eqn:gumbel_softmax}), we use an exponentially decaying temperature. After the search finishes, we sample several architectures from the trained distribution $P_{\bm{\theta}}$, and train them from scratch. Our architecture search framework is implemented in pytorch \cite{paszke2017automatic} and searched models are trained in Caffe2. More training details will be provided in the supplementary materials. 

Our experiment results are summarized in Table \ref{tab:imagenet}. We compare our searched models with state-of-the-art efficient models both designed automatically and manually. The primary metrics we care about are top-1 accuracy on the ImageNet validation set and the latency. If the latency is not available, we use FLOP as the secondary efficiency metric. For baseline models, we directly cite the parameter size, FLOP count, and top-1 accuracy from the original paper. Since our network is deployed with caffe2 with highly efficient in8 implementation, we have an unfair latency advantage against other baselines. Therefore, we implement the baseline models ourselves and measure their latency under the same environment for a fair comparison. For automatically designed models, we also compare the search method, search space, and search cost. 

\begin{table*}[h]
\centering
\begin{tabular}{c|ccc|cccc}
\hline
Model                                         & \begin{tabular}[c]{@{}c@{}}Search \\ method\end{tabular} & \begin{tabular}[c]{@{}c@{}}Search \\ space\end{tabular} & \begin{tabular}[c]{@{}c@{}}Search cost\\ (GPU hours / relative)\end{tabular} & \#Params & \#FLOPs & \begin{tabular}[c]{@{}c@{}}CPU \\ Latency\end{tabular} & \begin{tabular}[c]{@{}c@{}}Top-1 \\ acc (\%)\end{tabular} \\ \hline
1.0-MobileNetV2 \cite{sandler2018mobilenetv2} & manual                                                   & -                                                       & -                                                               & 3.4M         & 300M    & 21.7 ms                                                & 72.0                                                      \\
1.5-ShuffleNetV2 \cite{ma2018shufflenet}      & manual                                                   & -                                                       & -                                                               & 3.5M         & 299M    & 22.0 ms                                                & 72.6                                                      \\
CondenseNet (G=C=8) \cite{huang2017condensenet}& manual                                                   & -                                                       & -                                                               & 2.9M         & 274M    & 28.4 $^\ddag$ ms                                                & 71.0                                                      \\
MnasNet-65 \cite{ma2018shufflenet}            & RL                                                       & stage-wise                                              & 91K$^*$ / 421x                                     & 3.6M         & 270M    & -                                                      & 73.0                                                     \\ 
DARTS  \cite{liu2018darts}                    & gradient                                                 & cell                                                    & 288 / 1.33x                                                    & 4.9M         & 595M    & -                                                      & \textbf{73.1}                                                      \\ 
FBNet-A (ours)                              & gradient                                                 & layer-wise                                              & 216 / 1.0x                                                    & 4.3M         & \textbf{249M}    & \textbf{19.8 ms}                                                & 73.0                                                      \\ \hline
1.3-MobileNetV2 \cite{sandler2018mobilenetv2} & manual                                                   & -                                                       & -                                                               & 5.3M         & 509M    & 33.8 ms                                                 & \textbf{74.4}                                                      \\
CondenseNet (G=C=4) \cite{huang2017condensenet}& manual                                                   & -                                                       & -                                                               & 4.8M         & 529M    & 28.7$^\ddag$ ms                                               & 73.8                                                      \\
MnasNet \cite{tan2018mnasnet}                 & RL                                                       & stage-wise                                              & 91K$^*$ / 421x                                      & 4.2M         & 317M    & 23.7 ms                                                & 74.0                                                      \\
NASNet-A  \cite{zoph2017learning}             & RL                                                       & cell                                                    & 48K / 222x                                       & 5.3M         & 564M    & -                                                      & 74.0                                                      \\
PNASNet \cite{liu2017progressive}             & SMBO                                                     & cell                                                    & 6K$^{\dag}$ / 27.8x                                      & 5.1M         & 588M    & -                                                      & 74.2                                                      \\ 
FBNet-B (ours)                              & gradient                                                 & layer-wise                                              & 216 / 1.0x                                                     & 4.5M         & \textbf{295M}    & \textbf{23.1 ms}                                                & 74.1                                                      \\ \hline
1.4-MobileNetV2 \cite{sandler2018mobilenetv2} & manual                                                   & -                                                       & -                                                              & 6.9M         & 585M    & 37.4 ms                                                   & 74.7                                                      \\
2.0-ShuffleNetV2 \cite{ma2018shufflenet}      & manual                                                   & -                                                       & -                                                              & 7.4M         & 591M    & 33.3 ms                                                   & \textbf{74.9}                                                      \\
MnasNet-92 \cite{tan2018mnasnet}              & RL                                                       & stage-wise                                              & 91K$^*$ / 421x                                         & 4.4M         & 388M    & -                                                      & 74.8                                                     \\
FBNet-C (ours)                              & gradient                                                 & layer-wise                                              & 216 / 1.0x                                                       & 5.5M         & \textbf{375M}    & \textbf{28.1 ms}                                                   & \textbf{74.9}                                                     \\ \hline
\end{tabular}
\caption{ImageNet classification performance compared with baselines. For baseline models, we directly cite the parameter size, FLOP count and top-1 accuracy on the ImageNet validation set from their original papers. For CPU latency, we deploy and benchmark the models on the same Samsung Galaxy S8 phone with Caffe2 int8 implementation. The details of MnasNet-\{64, 92\} are not disclosed from \cite{tan2018mnasnet} so we cannot measure the latency. *The search cost for MnasNet is estimated according to the description in \cite{tan2018mnasnet}. $\dag$ The search cost is estimated based on the claim from \cite{liu2017progressive} that PNAS \cite{liu2017progressive} is 8x lower than NAS\cite{zoph2017learning}. $\ddag$ The inference engine is faster than other models. } 
\label{tab:imagenet}
\end{table*}

Table \ref{tab:imagenet} divides the models into three categories according to their accuracy level. In the first group, FBNet-A achieves 73.0\% accuracy, better than 1.0-MobileNetV2 (+1.0\%), 1.5-ShuffleNet V2 (+0.4\%), and CondenseNet (+2\%), and are on par with DARTS and MnasNet-65. Regarding latency, FBNet-A is 1.9 ms (relative 9.6\%), 2.2 ms (relative 11\%), and 8.6 ms (relative 43\%) better than the MobileNetV2, ShuffleNetV2, and CondenseNet counterparts. Although we did not optimize for FLOP count directly, FBNet-A's FLOP count is only 249M, 50M smaller (relative 20\%) than MobileNetV2 and ShuffleNetV2, 20M (relative 8\%) smaller than MnasNet, and 2.4X smaller than DARTS. In the second group, FBNet-B achieves comparable accuracy with 1.3-MobileNetV2, but the latency is 1.46x lower, and the FLOP count is 1.73x smaller, even smaller than 1.0-MobileNetV2 and 1.5-ShuffleNet V2. Compared with MnasNet, FBNet-B's accuracy is 0.1\% higher, latency is 0.6ms lower, and FLOP count is 22M (relative 7\%) smaller. We do not have the latency of NASNet-A and PNASNet, but the accuracy is comparable, and the FLOP count is 1.9x and 2.0x smaller. In the third group, FBNet-C achieves 74.9\% accuracy, same as 2.0-ShuffleNetV2 and better than all others. The latency is 28.1 ms, 1.33x and 1.19x faster than MobileNet and ShuffleNet. The FLOP count is 1.56x, 1.58x, and 1.03x smaller than MobileNet, ShuffleNet, and MnasNet-92.

Among all the automatically searched models, FBNet's performance is much stronger than DARTS, PNAS, and NAS, and better than MnasNet. However, the search cost is orders of magnitude lower. MnasNet \cite{tan2018mnasnet} does not disclose the exact search cost (in terms of GPU-hours). However, it mentions that the controller samples 8,000 models during the search and each model is trained for five epochs. According to our experiments, training of MNasNet for one epoch takes 17 minutes using 8 GPUs. So the estimated cost for training 8,000 models for 5 epochs is about $17/60\times 5 \times 8 \times 8,000 \approx 91\times 10^3$ GPU hours. In comparison, the FBNet search takes 8 GPUs for only 27 hours, so the computational cost is only 216 GPU hours, or 421x faster than MnasNet, 222x faster than NAS, 27.8x faster than PNAS, and 1.33x faster than DARTS.

\begin{figure}[h]
\begin{center}
\includegraphics[width=1.\linewidth]{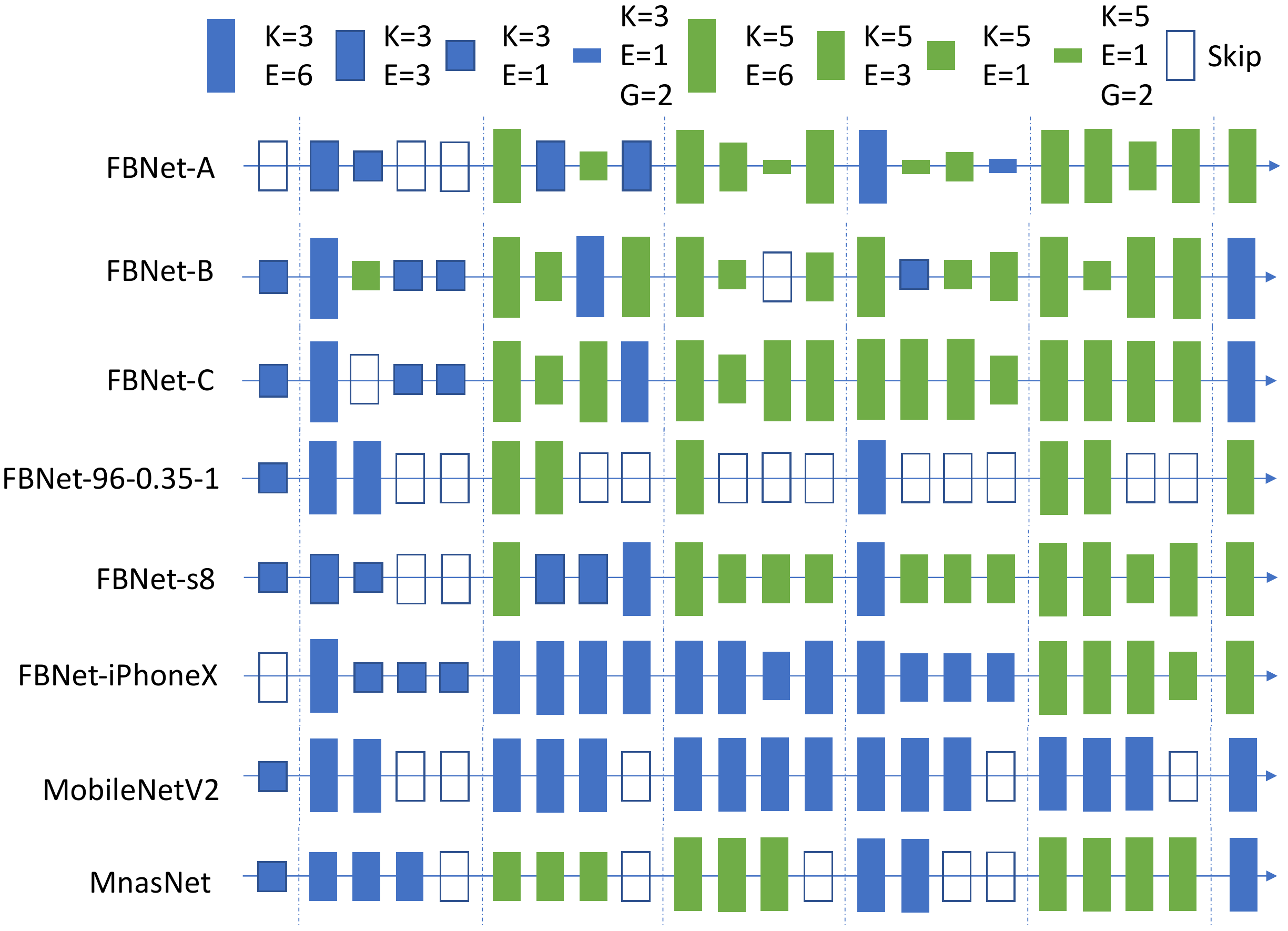}
\end{center}
\caption{Visualization of some of the searched architectures. We use rectangle boxes to denote blocks for each layer. We use different colors to denote the kernel size of the depthwise convolution, blue for kernel size of 3, green for kernel size of 5, and empty for skipping. We use height to denote the expansion rate of the block: 6, 3, 1, and 1 with group-2 convolution.}
\label{fig:arch_viz}
\vspace{-10pt}
\end{figure}

We visualize some of our searched FBNets, MobileNetV2, and MnasNet in Figure \ref{fig:arch_viz}.

\subsection{Different Resolution and Channel Size Scaling}
A common technique to reduce the computational cost of a ConvNet is to reduce the input resolution or channel size without changing the ConvNet structure. This approach is likely to be sub-optimal. We hypothesize that with a different input resolution and channel size scaling, the optimal ConvNet structure will be different. To test this, we use DNAS to search for several different combinations of input resolution and channel size scaling. Thanks to the superior efficiency of DNAS, we can finish the search very quickly. The result is summarized in Table \ref{tab:scaling}. Compared with MobileNetV2 under the same input size and channel size scaling, our searched models achieve 1.5\% to 6.4\% better accuracy with similar latency. Especially the FBNet-96-0.35-1 model achieves 50.2\% (+4.7\%) accuracy and 2.9 ms latency (345 frames per second) on a Samsung Galaxy S8.

\begin{table*}[]
\centering
\begin{tabular}{c|c|cccc}
\hline
\begin{tabular}[c]{@{}c@{}}Input size \&\\ Channel Scaling\end{tabular} & Model                  & \#Parameters & \#FLOPs & CPU Latency & Top-1 acc (\%) \\ \hline
\multirow{3}{*}{(224, 0.35)}                                            & MobileNetV2-224-0.35   & 1.7M         & 59M     & 9.3 ms      & 60.3           \\
                                                                        & MNasNet-scale-224-0.35 & 1.9M         & 76M     & 10.7 ms     & 62.4 (+2.1)    \\
                                                                        & FBNet-224-0.35          & 2.0M         & 72M     & 10.7 ms     & 65.3 (+5.0)    \\ \hline
\multirow{3}{*}{(192, 0.50)}                                            & MobileNetV2            & 2.0M        & 71M     & 8.4 ms      & 63.9           \\
                                                                        & MnasNet-search-192-0.5 & -            & -       & -           & 65.6 (+1.7)    \\
                                                                        & FBNet-192-0.5 (ours)    & 2.6M        & 73M     & 9.9 ms      & 65.9 (+2.0)    \\ \hline
\multirow{3}{*}{(128, 1.0)}                                             & MobileNetV2            & 3.5M         & 99M     & 8.4 ms      & 65.3           \\
                                                                        & MnasNet-scale-128-1.0  & 4.2M         & 103M    & 9.2 ms      & 67.3 (+2.0)    \\
                                                                        & FBNet-128-1.0 (ours)    & 4.2M        & 92M     & 9.0 ms      & 67.0 (+1.7)    \\ \hline
\multirow{2}{*}{(128, 0.50)}                                            & MobileNetV2            & 2.0M        & 32M     & 4.8 ms      & 57.7           \\
                                                                        & FBNet-128-0.5 (ours)    & 2.4M        & 32M   & 5.1 ms      & 60.0 (+2.3)    \\ \hline
\multirow{3}{*}{(96, 0.35)}                                             & MobileNetV2            & 1.7M        & 11M     & 3.8 ms      & 45.5           \\
                                                                        & FBNet-96-0.35-1 (ours)  & 1.8M        & 12.9M   & 2.9 ms      & 50.2 (+4.7)    \\
                                                                        & FBNet-96-0.35-2 (ours)  & 1.9M        & 13.7M   & 3.6 ms      & 51.9 (+6.4)    \\ \hline
\end{tabular}
\caption{FBNets searched for different input resolution and channel scaling. MnasNet-scale is the MnasNet model with input and channel size scaling. MnasNet-search-192-0.5 is a model searched with an input size of 192 and channel scaling of 0.5. Details of it are not disclosed in \cite{tan2018mnasnet}, so we only cite the accuracy.}
\label{tab:scaling}
\end{table*}

\begin{table*}[h]
\centering
\begin{tabular}{c|ccccc}
\hline
Model & \#Parameters & \#FLOPs & \begin{tabular}[c]{@{}c@{}}Latency on \\ iPhone X\end{tabular} & \begin{tabular}[c]{@{}c@{}}Latency on\\ Samsung S8\end{tabular} & Top-1 acc (\%) \\ \hline
FBNet-iPhoneX      & 4.47M        & 322M    & 19.84 ms (target)                                                      & 23.33 ms                                                        & 73.20         \\
FBNet-S8           & 4.43M        & 293M    & 27.53 ms                                                               & 22.12 ms (target)                                               & 73.27         \\ \hline
\end{tabular}
\caption{FBNets searched for different devices.}
\label{tab:device}
\end{table*}

We visualize the architecture of FBNet-96-0.35-1 in Figure \ref{fig:arch_viz}, we can see that many layers are skipped, and the network is much shallower than FBNet-\{A, B, C\}, whose input size is 224. We conjecture that this is because with smaller input size, the receptive field needed to parse the image also becomes smaller, so having more layers will not effectively increase the accuracy.

\subsection{Different Target Devices}
In previous ConvNet design practices, the same ConvNet model is deployed to many different devices. However, this is sub-optimal since different computing platforms and software implementation can have different characteristics. To validate this, we conduct search targeting two mobile devices: Samsung Galaxy S8 with Qualcomm Snapdragon 835 platforms, and iPhone X with A11 Bionic processors. We use the same architecture search space, but different latency lookup tables collected from two target devices. All the architecture search and training protocols are the same. After we searched and trained two models, we deploy them to both Samsung Galaxy S8 and iPhone X to benchmark the overall latency. The result is summarized in Table. \ref{tab:device}. 

As we can see, the two models reach similar accuracy ($73.20\%$ vs. $73.27\%$). FBNet-iphoneX model's latency is 19.84 ms on its target device, but when deployed to a Samsung S8, its latency increases to 23.33 ms. On the other hand, FBNet-S8 reaches a latency of 22.12 ms on a Samsung S8, but when deployed to an iPhone X, the latency hikes to 27.53 ms, 7.69 ms (relatively 39\%) higher than FBNet-iPhone X. 
This demonstrates the necessity of re-designing ConvNets for different target devices. 

\begin{figure}[h]
\begin{center}
\includegraphics[width=.85\linewidth]{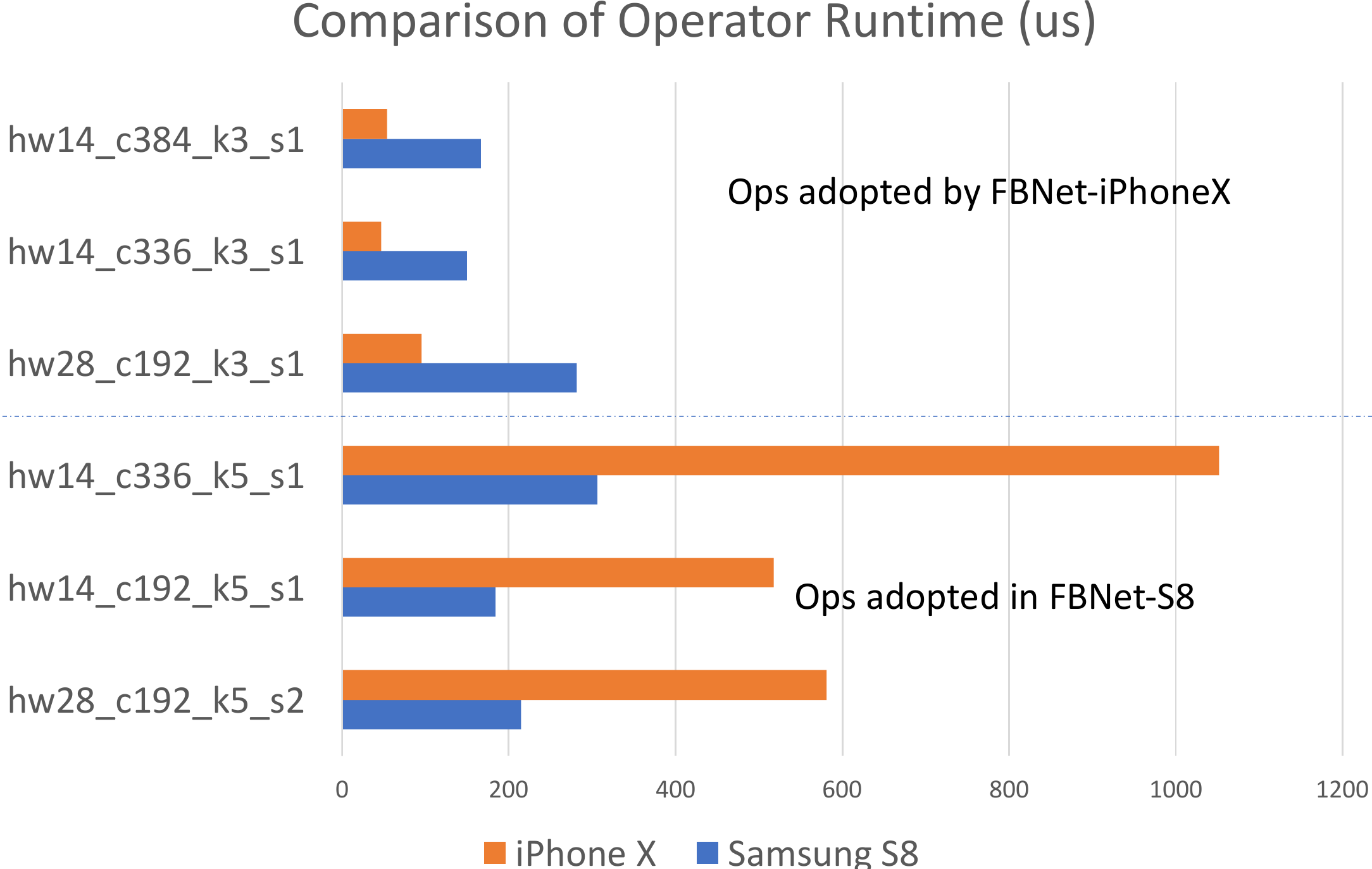}
\end{center}
\caption{Comparison of operator runtime on two devices. Runtime is in micro-second (us). Orange bar denotes the runtime on iPhone X and blue bar denotes the runtime on Samsung S8. The upper three operators are faster on iPhone X, therefore they are automatically adopted in FBNet-iPhoneX. The lower three operators are faster on Samsung S8, and they are also automatically adopted in FBNet-S8.}
\label{fig:runtime_comparision}
\vspace{-14pt}
\end{figure}

Two models are visualized in Figure \ref{fig:arch_viz}. Note that FBNet-S8 uses many blocks with 5x5 depthwise convolution while FBNet-iPhoneX only uses them in the last two stages. We examine the depthwise convolution operators used in the two models and compare their runtime on both devices. As shown in Figure \ref{fig:runtime_comparision}, the upper three operators are faster on iPhone X, therefore they are automatically adopted in FBNet-iPhoneX. The lower three operators are significantly faster on Samsung S8, and they are also automatically adopted in FBNet-S8. Notice the drastic runtime differences of the lower three operators on two target devices. It explains why the Samsung-S8-optimized model performs poorly on an iPhone X. This shows DNAS can automatically optimize the operator adoptions and generate different ConvNets optimized for different devices. 

\section{Conclusion}
We present DNAS, a differentiable neural architecture search framework. It optimizes over a layer-wise search space and represents the search space by a stochastic super net. The actual target device latency of blocks is used to compute the loss for super net training. FBNets, a family of models discovered by DNAS surpass state-of-the-art models, both manually and automatically designed: FBNet-B achieves 74.1\% top-1 accuracy with 295M FLOPs and 23.1 ms latency, 2.4x smaller and 1.5x faster than MobileNetV2-1.3 with the same accuracy. It also achieves better accuracy and lower latency than MnasNet, the state-of-the-art efficient model designed automatically; we estimate the search cost of DNAS is 420x smaller. Such efficiency allows us to conduct searches for different input resolutions and channel scaling. Discovered models achieve 1.5\% to 6.4\% accuracy gains. The smallest FBNet achieves 50.2\% accuracy with a latency of 2.9 ms (345 frames/sec) with batch size 1. Over the Samsung-optimized FBNet, the improved FBNet achieves 1.4x speed up on an iPhone X, showing DNAS is able to adapt to different target devices automatically. 
\clearpage
{\small
\bibliographystyle{ieee}
\bibliography{egbib}
}

\clearpage
\appendix
\section{Experiment details}
We describe more experiment details in this appendix to facilitate other researchers to reproduce our work. Our architecture search is divided into two stages. In the first stage, we train the stochastic super net to find an optimal architecture distribution. In the second stage, we sample architectures from the distribution and train them from scratch. 

To train the stochastic super net, we randomly sample 100 classes from the original 1,000 classes of ImageNet. Training the super net on this smaller proxy dataset is much faster. We train the stochastic super net for 90 epochs with a batch size of 192. In each epoch, we first train the operator parameters $w_a$ on 80\% of the training set using stochastic gradient descent with momentum. The initial learning rate is 0.1, and decay following a cosine decaying schedule. The momentum is 0.9, and weight decay is $10^{-4}$. Next, we train the architecture distribution parameter $\bm{\theta}$ on the rest 20\% of the training set with Adam optimizer \cite{kingma2014adam} with a learning rate of $10^{-2}$ and weight decay of $5\times 10^{-4}$. The split of weight and architecture parameter training ensure the architecture generalize to the validation dataset. To control the Gumbel Softmax in (\ref{eqn:gumbel_softmax}), we use an initial temperature of 5.0 and exponentially anneal it by $exp(-0.045)\approx 0.956$ every epoch. For the loss function in (\ref{eqn:loss}), we set $\alpha$ to 0.2 and $\beta$ to 0.6. We use the standard ResNet data augmentation \cite{he2016deep} to process the input images. We found that at the beginning of the training, operators are usually not sufficiently trained, so their contributions to the accuracy are not clear. However, their computational costs are always significantly different from each other. As a consequence, the super net may always pick low computational cost operators at the beginning of the training. To prevent this, we postpone the training of the architecture parameter $\bm{\theta}$ by 10 epochs to allow operator weights to be sufficiently trained first. At the end of the super net training, we sample 6 architectures from the final distribution to be trained from scratch.

To train the sampled architectures, the training protocols are different for different models. Here we describe the training protocol for FBNet-\{A, B, C\}. These models have an input resolution of 224, channel size scaling of 1.0. We train the models with a batch size of 256 on 8 GPUs for 360 epochs. We set the initial learning rate to be 0.1, and decay 10x at 90, 180, and 270 epochs. The momentum is 0.9, weight decay is $4\times 10^{-5}$. We use dropout at the last convolution layer of the network, and the dropout ratio is 0.2. We use the standard GoogleNet data augmentation \cite{szegedy2015going} to randomly resize the image during training. 

\end{document}